\documentclass{article}

\usepackage[preprint]{neurips_2026}

\usepackage[utf8]{inputenc} % allow utf-8 input
\usepackage[T1]{fontenc}    % use 8-bit T1 fonts
\usepackage{hyperref}       % hyperlinks
\usepackage{url}            % simple URL typesetting
\usepackage{booktabs}       % professional-quality tables
\usepackage{amsfonts}       % blackboard math symbols
\usepackage{nicefrac}       % compact symbols for 1/2, etc.
\usepackage{microtype}      % microtypography
\usepackage{xcolor}         % colors
\usepackage{amsmath} 
\usepackage{amssymb}
\usepackage{booktabs} 
\usepackage{multirow}
\usepackage{graphicx}
\usepackage{booktabs} % For the results table
\usepackage{caption}
\usepackage{multido}
\usepackage{blindtext}
\usepackage[compact]{titlesec}
\titlespacing{\section}{0pt}{2ex}{1ex}
\titlespacing{\subsection}{0pt}{1ex}{0.5ex}

\title{GeoStack: A Framework for Quasi-Abelian Knowledge Composition in VLMs}
\author{%
  Pranav Mantini \\
  Department of Computer Science\\
  University of Houston\\
  \texttt{pmantini@uh.edu} \\
  \And
  Shishir K.~Shah \\
  The University of Oklahoma \\
  School of Computer Science \\
  \texttt{sshah@ou.edu} \\
}

\begin{document}

\maketitle

\begin{abstract}
We address the challenge of knowledge composition in Vision-Language Models (VLMs), where accumulating expertise across multiple domains or tasks typically leads to catastrophic forgetting. We introduce GeoStack (Geometric Stacking), a modular framework that allows independently trained domain experts to be composed into a unified model. By imposing geometric and structural constraints on the adapter manifold, GeoStack ensures the foundational knowledge of the base model is preserved. Furthermore, we mathematically demonstrate a weight-folding property that achieves constant-time inference complexity ($O(1)$), regardless of the number of integrated experts. Experimental results across multi-domain adaptation and class-incremental learning show that GeoStack provides an efficient mechanism for long-term knowledge composition while significantly mitigating catastrophic forgetting. Code is available at~\url{https://github.com/QuantitativeImagingLaboratory/GeoStack}.
\end{abstract}    
\section{Introduction and Motivating Work}

Single-task models, such as classifiers, gather knowledge to achieve a domain-specific objective. In contrast, multitask learning~\citep{caruana1997multitask} aims to incorporate knowledge from multiple objectives for broader applicability. Similarly, incremental learning aims to expand a model's capabilities on novel data, typically within the same domain. At their core, these problems aim to achieve \textbf{knowledge composition}.
% , in which knowledge from multiple domains is either integrated into a single model for multi-task applicability or existing knowledge is combined with novel information to expand the model's capabilities.

Multi-Task Learning~\citep{LiuQMTL} involves joint training strategies on disparate data distributions and quickly becomes infeasible as the number of tasks increases. Furthermore, this approach increases model complexity and requires addressing complex challenges such as class imbalance and hyperparameter selection. Sequential fine-tuning is a viable alternative, where an existing model is fine-tuned with novel data. However, these models are prone to catastrophic forgetting~\citep{kirkpatrick2017overcoming}.   

\textbf{Catastrophic forgetting} is a phenomenon where connectionist models, when fine-tuned on new data, fail to retain their original knowledge. Classic approaches to address this include Knowledge Distillation~\citep{hinton2015distilling, li2017learning}, which regularizes student predictions against a frozen teacher, and Data Replay~\citep{icarlRebuffi}, which interleaves exemplary data from previous tasks with data from new tasks to maintain consistency.

\textbf{Adapter-based} methods have emerged as a flexible alternative for multitask learning. In this paradigm, models share a large frozen backbone and train a small set of parameters for domain-specific tasks. ~\cite{houlsby2019parameter} has trained single-task adapters for text classification tasks, and ~\cite{SticklandBERT} have proposed multitask adapters for the Recognizing Textual Entailment dataset that match the performance of fine-tuned models.   
% Single-task adapters can be trained to achieve optimal performance on specific tasks~\cite{houlsby2019parameter}, while multitask adapters can be optimized with a combined objective across several domains~\cite{SticklandBERT}.

Single-task adapters generally do not allow for the sharing of information between tasks. To overcome this, ~\cite{pfeiffer2021adapterfusion} proposed a two-stage mechanism: (1) \textbf{Knowledge Extraction}, where task-specific representations are encapsulated within adapters, and (2) \textbf{Knowledge Composition}, which employs a fusion mechanism to combine expertise across adapters for multitask scenarios. ~\cite{chaichana2025decomrenormmergemodelmergingright} proposed Decom-Renorm-Merge that utilizes Singular Value Decomposition (SVD) and renormalization to fuse independently trained checkpoints into a single multitasking model. \textbf{Task Arithmetic} proposed by ~\cite{ilharco2023editingmodelstaskarithmetic} further simplifies this by representing each task as a \textbf{task vector} (the difference between fine-tuned and pre-trained weights), which can be linearly summed to merge multiple capabilities into a single model without additional parameters. Furthermore, benchmarks like VL-Adapter~\citep{sung2022vladapterparameterefficienttransferlearning} demonstrate that weight-sharing mechanisms across vanilla adapters can often match the performance of full fine-tuning with significantly less parameter overhead. Such methods aim to learn expertise independently, without the need for simultaneous data access or prohibitive retraining.

% ; however, they often struggle to maintain order-invariance or constant-time inference as the number of tasks scales.

\textit{Knowledge Extraction using VLMs:}
VLM-based adapters, particularly those utilizing Contrastive Language-Image Pre-training (CLIP~\citep{radford2021learningtransferablevisualmodels}) as a backbone, have proven to be an efficient mechanism for rapid domain adaptation. These generally fall into two paradigms:
1) Prompt-based Tuning: Methods such as CoOp~\cite{Zhou_2022} and Co-CoOp~\citep{zhou2022conditionalpromptlearningvisionlanguage} optimize learnable prompt vectors to adapt a model for a targeted domain distribution.
2) Adapter-based Tuning: Techniques like CLIP-Adapter~\citep{gao2025clipadapterbettervisionlanguagemodels} and Tip-Adapter~\citep{zhang2021tipadaptertrainingfreeclipadapterbetter} introduce lightweight bottleneck layers to refine features for specific domains.

Despite these advancements, knowledge composition mechanisms for VLM adapters remain largely nonexistent. We advocate that an ideal framework for knowledge composition must satisfy the following fundamental principles:
\begin{enumerate}
    \item \textit{Independent Training:} Adapters must be trainable independently without the need for cross-domain or historical data, and the need for joint-training hyperparameters.
    \item \textit{Modularity:} Adapters should be modular, allowing the integration of knowledge without re-training the ensemble.
    \item \textit{Order-Invariance:} The composed knowledge should be invariant to the order of integration, thus removing the need for combinatorial optimization.
    \item \textit{Foundational Preservation:} Composition must not degrade the model’s original capabilities or disrupt the foundational knowledge.
    \item \textit{Computational Efficiency:} Architectural complexity should remain constant or, at most, grow linearly with each added task.
\end{enumerate}

Recent studies have enabled a deeper understanding of CLIP’s geometric properties, specifically the well-known modality gap~\citep{modalitygap} and the canonical relations~\citep{gupta2026canonicalizingmultimodalcontrastiverepresentation} observed between the feature distributions of independently trained VLMs. Building on these insights, Bilinear CLIP (BiCLIP)~\citep{mantini2026biclipdomaincanonicalizationstructured} proposes domain canonicalization for few-shot adaptation by introducing a learnable geometric transformation matrix $W$. While zero-shot CLIP computes classification probabilities via the dot product $IT^\top$ between image ($I$) and text ($T$) features, BiCLIP optimizes the transformed product $IWT^\top$.

\begin{figure}[th]
  % 1. Use slightly less than 0.5 to allow for a small gap
  \begin{minipage}{0.54\textwidth}
    \centering
    \includegraphics[width=\textwidth]{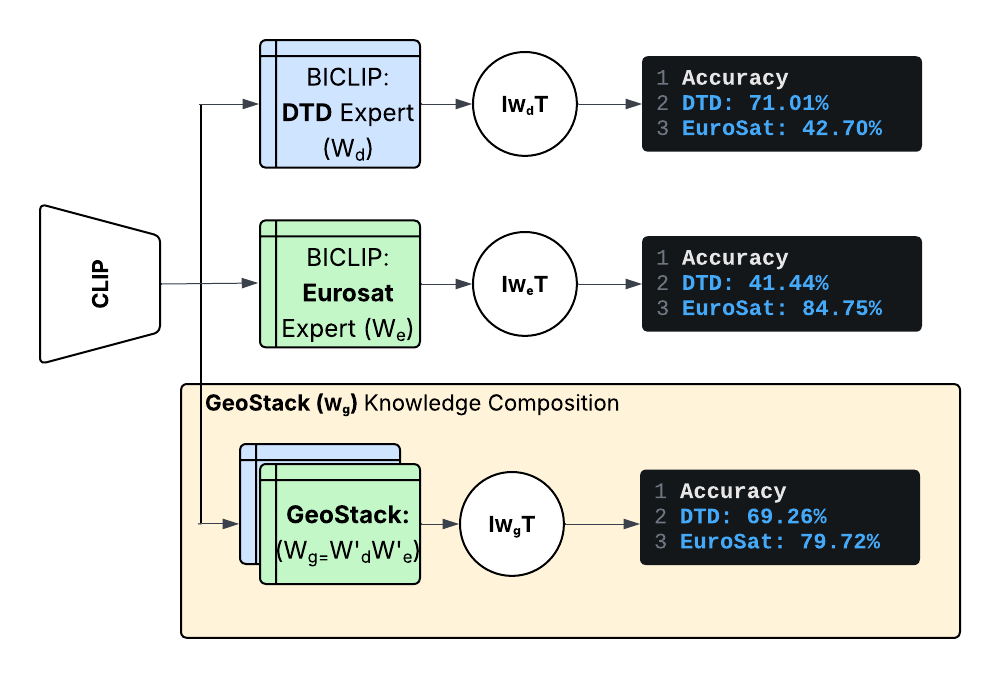}
  \end{minipage}
  \hfill % 2. This pushes the two minipages to the far edges
  \begin{minipage}{0.44\textwidth}   
    % 3. Put the caption here. It will wrap within this 0.48 width.
    \caption{\textbf{GeoStack Overview.} Domain-specific adapters (BiCLIP) are not applicable across domains and have reduced generalizability. GeoStack allows domain experts ($W'_d, W'_e$) to be trained independently and then stacked to enable applicability across multiple tasks.}
    \label{fig:geostack_intro_concept}
  \end{minipage}
\end{figure}
BiCLIP is an efficient and geometrically interpretable mechanism for domain adaptation. However, these transformations are domain-specific. As shown in Figure~\ref{fig:geostack_intro_concept}, a BiCLIP expert optimized for the DTD~\citep{dtd} domain achieves $71.01\%$ accuracy on its target but falls to $42.70\%$ on the EuroSAT~\citep{eurosat} dataset. Conversely, a EuroSAT expert achieves $84.75\%$ on its own domain but drops to $41.44\%$ on DTD.

We propose \textbf{Geometric Stacking (GeoStack)}, a modular knowledge composition framework designed to aggregate expertise from multiple domain-specific adapters into a multi-expert model with zero additional inference complexity. Specifically, BiCLIP adapters are trained with geometric constraints to produce domain-specific Geometric Layers (GeoLayers) that can be stacked onto one another for multitask performance. As shown in Figure~\ref{fig:geostack_intro_concept}, GeoStack performs knowledge composition to maintain the performance across both DTD ($69.26\%$) and Eurosat ($79.72\%$) datasets. This approach allows an arbitrary number of experts to be composed via matrix multiplication and folded into a single weight matrix, maintaining $O(1)$ complexity regardless of the number of domains.

% It offers three distinct advantages over traditional composition methods:

% \textit{Decoupled Training:} Unlike distillation or replay-based strategies, GeoStack experts are trained in total isolation. Each adapter ($W_i$) is learned using only its local domain data, eliminating the requirement for simultaneous access to multiple datasets or the tuning of complex joint-training hyperparameters.

% \textit{Modularity and Abelian Nature:} GeoStack is inherently modular, enabling the sequential stacking of expertise. The adapters exhibit an Abelian (commutative) property, allowing the experts to be integrated in any arbitrary order, removing the need for expensive combinatorial searches to find an optimal stacking configuration.

% \textit{Zero Inference Overhead:}  GeoStack employs a weight-folding mechanism that seamlessly integrates composed knowledge into the original CLIP architecture. The parameter count and computational complexity of a GeoStack model—regardless of the number of domains integrated—remain identical to the original CLIP foundation.

Our primary contributions are as follows:
\begin{itemize}
\item \textbf{The GeoStack Framework}: We introduce GeoStack, a modular framework for knowledge composition in VLMs. We derive the geometric constraints necessary to ensure the stability of the framework.
\item \textbf{Theoretical Foundations of Stackability}: We define the metrics and conditions under which domain-specific experts can be stably composed using GeoStack.
\item \textbf{The Weight-Folding Property}: We demonstrate that GeoStack adapters enable multitask inference with constant-time complexity ($O(1)$), independent of the number of experts used in the composition.
\item \textbf{Empirical Validation}: We conduct extensive experiments across multi-domain adaptation and class-incremental learning, demonstrating GeoStack's superior performance and its resistance to catastrophic forgetting.
\end{itemize}

\section{GeoStack Theory}
\label{sec:theory}
% \subsection{Preliminaries: Zero-shot CLIP classification}
\subsection{Preliminaries}
CLIP projects images and textual prompts into a shared embedding space $\mathbb{R}^d$, yielding features $I$ and $T$. In the zero-shot setting, the similarity (dot product) between these representations is used to compute the posterior for classification. Given a matching positive pair $(I, T^+)$ and an unmatched negative pair $(I, T^-)$, the classification decision boundary and the resulting margin $M_z$ are expressed as:
\begin{equation}
I {T^+}^\top > I {T^-}^\top \implies M_z = I {T^+}^\top - I {T^-}^\top > 0
\end{equation}

However, CLIP is trained on generic web data, and this pre-trained geometric boundary is often inadequate in specialized domains.

\textbf{BilinearCLIP}~\citep{mantini2026biclipdomaincanonicalizationstructured} (BiCLIP) hypothesizes that the domain-specific decision boundary can be recovered by applying a geometric transformation $W \in \mathbb{R}^{d \times d}$ to the image features ($I'=IW$). For a domain $\mathcal{D}_a$, the BiCLIP margin $M_a$ is defined as:
\begin{equation}
M_a = I_a W_a {T_a^+}^\top - I_a W_a {T_a^-}^\top > 0
\end{equation}
While $W_a$  optimizes the decision boundary for domain $\mathcal{D}_a$, it degrades the generalization capabilities of CLIP, resulting in a model that is not applicable to other domains. 

\subsection{GeoStack: Problem Formulation}
Building on BiCLIP, GeoStack is a modular architecture that allows the composition of multiple experts via sequential matrix multiplication. The composite operator for domains $\mathcal{D}_a$ and $\mathcal{D}_b$ is defined as $W_g = W_a W_b$. The primary challenge is in ensuring that the subsequent expert $W_b$ does not destroy the margin $M_a$ previously established for $\mathcal{D}_a$. To ensure framework viability, the composite margin $M_g$ must satisfy:
\begin{equation}
I_a W_g {T_a^+}^\top - I_a W_g {T_a^-}^\top > 0 \quad \text{and} \quad I_b W_g {T_b^+}^\top - I_b W_g {T_b^-}^\top > 0
\end{equation}
For these inequalities to hold, the original margin $M_a$ must remain positive under the influence of subsequent operators. We require the framework to satisfy the stability condition: $M_a > 0 \implies M_g > 0$.
% For two domains $\mathcal{D}_a$ and $\mathcal{D}_b$: $W_{total} = \prod_{i=1}^N W_i$. The primary challenge in such a composition is ensuring that an expert learned for domain $\mathcal{D}_b$ does not destroy the decision boundary previously established for domain $\mathcal{D}_a$. We define the composed operator for domains $a$ and $b$ as $W_g = W_a W_b$. To ensure simultaneous operability, the margin $M_g$ for the composite operator must satisfy:

\subsection{GeoStack: Geometric Constraints for Multi-Domain Composition}
GeoStack ensures this margin stability by imposing two geometric and structural constraints from BiCLIP:

\textbf{Upper Triangular Closure:} We restrict each adapter $W$ to the set of upper-triangular matrices $\mathcal{U}$. Because $\mathcal{U}$ is closed under multiplication, any composed operator $W_{total} = W_a W_b \dots W_n$ remains upper-triangular, ensuring the composite operator is always a valid member of the same transformation class.

\textbf{Perturbation Prior:} Each learnable adapter $W$ is initialized as the identity matrix $\mathbf{I} \in \mathcal{U}$. This initialization acts as a geometric prior. Consequently, the learned transformation can be viewed as a perturbation $W = \mathbf{I} + \Delta$, where $\Delta \in \mathcal{U}$ represents the domain-specific geometric shift. 
    
    % However, while BiCLIP allows $\Delta$ to be unconstrained, GeoStack requires these perturbations to be **minimal isometries** to allow for cross-domain operability. As we detail in Section 3.5, we enforce this "minimal rotation" via an orthogonality objective, ensuring the adapter acts as a near-isometric nudge localized around the identity rather than a global re-orientation.

\subsection{Perturbation Minimization Theory}
\label{sec:perturbation_minimization}
By defining each domain expert as a perturbation $W_i = \mathbf{I} + \Delta_i$, the GeoStack composition of two experts is given by:
\begin{equation}
W_a W_b = (\mathbf{I} + \Delta_a)(\mathbf{I} + \Delta_b) = \mathbf{I} + \Delta_a + \Delta_b + \Delta_a \Delta_b
\end{equation}
When the learned perturbations $\Delta_a$ and $\Delta_b$ are small, their product term becomes negligible ($\Delta_a\Delta_b \approx \mathbf{0}$) as a second-order effect. The composed margin for domain $\mathcal{D}_a$ becomes:
\begin{equation}
    M_g \approx I_a (\mathbf{I} + \Delta_a + \Delta_b) {T_a^+}^\top - I_a (\mathbf{I} + \Delta_a + \Delta_b) {T_a^-}^\top \approx M_a + \underbrace{\left[ I_a \Delta_b {T_a^+}^\top - I_a \Delta_b {T_a^-}^\top \right]}_{\epsilon}
\end{equation}

Here, $\epsilon$ represents the inter-domain interference caused by Domain $\mathcal{D}_b$. The stability of the composed margin relies inherently on the spectral norms of the perturbations ($\|\Delta_b\|_2 < \delta$) remaining small. This structural guarantee ensures that $\|\epsilon\| \ll M_a$. Therefore, the composed GeoStack margin $M_g$ remains positive as long as $M_a$ is sufficiently discriminative and the \textbf{spectral norm of the perturbations is minimized}. 

\subsection{Properties of GeoStack}
\label{sec:properties}
The perturbation minimization theory underpinning GeoStack yields highly desirable mathematical properties for knowledge composition.

\textbf{1. Quasi-Additive Composition:} The multiplicative composition of GeoStack effectively reduces to a quasi-additive operation: $W_{g} \approx \mathbf{I} + \sum_{i=1}^n \Delta_i$. This property ensures that as new domains are added, the foundational knowledge of CLIP and previously learned experts is preserved.

\textbf{2. Quasi-Abelian composition:} A direct corollary of the quasi-additive property is the commutativity of the GeoStack. As $W_a W_b \approx \mathbf{I} + \Delta_a + \Delta_b$, the order of composition becomes largely irrelevant ($W_a W_b \approx W_b W_a$). This grants the GeoStack framework a quasi-Abelian property, allowing for greater flexibility in knowledge composition.

\textbf{3. The Stacking Metric:} We utilize the normalized orthogonality error as a proxy for stacking compatibility. Substituting $W = \mathbf{I} + \Delta$, the error expands to $\|WW^\top - \mathbf{I}\|^2_F \approx \|\Delta + \Delta^\top\|^2_F$ (See Appendix~\ref{Appendix:A}). Since the Frobenius norm upper-bounds the spectral norm ($\|\Delta\|_F \geq \|\Delta\|_2$) and is more computationally efficient to calculate, it serves as a practical upper bound for the interference $\|\epsilon\|$. This yields a practical stacking metric: operators with high orthogonality error violate the condition $\|\epsilon\| < M_a$, subsequently leading to catastrophic forgetting.

\textbf{4. The Folding Trick ($O(1)$ Inference):} GeoStack enables Zero-Overhead Inference via weight folding. In CLIP, the visual projection head is a matrix $P \in \mathbb{R}^{d' \times d}$. Since each adapter $W_k$ is a $d \times d$ matrix, the entire stack can be pre-computed into a single effective projection matrix: $
% \begin{equation}
P_{eff} = P \cdot \prod_{k=0}^{n-1} W_{k}
% \end{equation}
$ 

This property ensures that the inference complexity is constant ($O(1)$) with respect to the number of tasks. Mathematically, $P_{eff}$ is structurally identical to the original vanilla CLIP projection, meaning GeoStack provides multi-domain expertise with zero additional latency or memory footprint during deployment.

\subsection{Limitations: Margin Erosion in Deep Stacking}
\label{sec:limitations}
\textbf{Margin Erosion:} The Quasi-Additive Property ($W_{g} \approx \mathbf{I} + \sum \Delta_i$) implies a linear accumulation of error. The total inter-domain interference for domain $\mathcal{D}_a$ is the sum of perturbations from all $n$ subsequent experts: $\epsilon_{total} = \sum_{i \neq a}^n \left[ I_a \Delta_i {T_a^+}^\top - I_a \Delta_i {T_a^-}^\top \right]$

As the stack deepens, the stability condition $\|\epsilon_{total}\| < M_a$ is eventually violated—a phenomenon we term \textbf{Margin Erosion}. This leads to a gradual degradation in domain-specific performance, eventually regressing to sub-optimal zero-shot CLIP performance or causing a manifold collapse as the error accumulates.

\section{GeoLayer}

A \textbf{Geometric Layer} (GeoLayer) is an evolution of the BiCLIP adapter, optimized with geometric constraints to enable knowledge composition. While a BiCLIP adapter ($W \in \mathbb{R}^{d \times d}$) is optimized with the objective of aligning image and text features within a single domain, it often disrupts foundational knowledge, resulting in catastrophic forgetting. In contrast, a GeoLayer is trained with a dual objective: (1) of achieving domain-specific alignment, while (2) satisfying geometric constraints to preserve previous knowledge. These GeoLayers can be stably composed into a \textbf{GeoStack} to enable a multi-domain expert without catastrophic forgetting.

\textbf{Alignment Objective:}
We utilize the InfoNCE contrastive loss for domain alignment. Given a batch of $N$ image-text feature pairs $(I_j, T_j)$ from domain $\mathcal{D}_i$, we compute the transformed image features $I'_j = I_j W_i$ and their corresponding text embeddings $T_j$. The alignment loss is defined as: 
% $
% % \begin{equation}
% \mathcal{L}_{align} = -\frac{1}{N} \sum_{j=1}^N \log \frac{\exp(\text{sim}(I_j W_i, T_j) / \tau)}{\sum_{k=1}^N \exp(\text{sim}(I_j W_i, T_k) / \tau)}
% % \end{equation}
% $
\begin{equation}
\mathcal{L}_{align} = -\frac{1}{2N} \sum_{j=1}^N \left[ \log \frac{e^{\text{sim}(\mathbf{I}'_j, \mathbf{T}j) / \tau}}{\sum_{k=1}^N e^{\text{sim}(\mathbf{I}'_j, \mathbf{T}_k) / \tau}} + \log \frac{e^{\text{sim}(\mathbf{I}'_j, \mathbf{T}j) / \tau}}{\sum_{k=1}^N e^{\text{sim}(\mathbf{I}'_k, \mathbf{T}_j) / \tau}} \right]
\end{equation}

where $\tau$ is the temperature parameter and $\text{sim}(\cdot)$ denotes cosine similarity. This objective allows the GeoLayer to learn a domain-specific transformation that aligns the image features with their corresponding text modality for classification.

% While this objective successfully recovers the decision boundary $M_i$, unconstrained optimization (as in the original BiCLIP) allows $W_i$ to adopt any geometry that satisfies the margin, often leading to large spectral norms and high orthogonality error.

\textbf{Stackability Objective:}
To ensure that each GeoLayer satisfies the stability requirements derived in Section~\ref{sec:theory}, we minimize the Frobenius norm of the deviation from orthogonality, which effectively bounds the spectral norm $\|\Delta_i\|_2$. We define the Orthogonality Loss as: $
% \begin{equation}
\mathcal{L}_{ortho} = \|W_i^\top W_i - \mathbf{I}\|_F^2
% \end{equation}
$.

This objective ensures that the learned perturbation $\Delta_i$ remains minimal. Furthermore, by enforcing $W_i$ to remain in the neighborhood of an orthogonal matrix ensures the transformation is near-isometric. This preserves the feature norms ($\|I W_i\| \approx \|I\|$) during both training and inference.

% , allowing multiple GeoLayers to be composed without interfering with the previous knowledge.

\textbf{Convex Orthogonality Alignment Loss:}
The final optimization objective for a GeoLayer is a convex combination of the alignment and stackability objectives. We define the \textbf{Convex Orthogonality Alignment (COA)} Loss as: 
\begin{equation}
\label{Eq:convex_ortho_loss}
\mathcal{L}_{COA} = (1 - \lambda) \mathcal{L}_{align} + \lambda \mathcal{L}_{ortho}
\end{equation}

This formulation enables a calibration of the GeoLayer's behavior. As $\lambda \to 0$, the objective prioritizes domain-specific alignment. Conversely, as $\lambda \to 1$, the objective prioritizes the stability requirement for knowledge composition.

% In our experiments, we demonstrate that a balanced $\lambda$ allows experts to achieve significant alignment gains while maintaining the geometric integrity required for deep stacking.

% \subsection{The Folding Trick: On the zero computation complexity overhead}

% \textbf{Decoupled Optimization:} A key advantage of the GeoStack framework is that each GeoLayer is trained in complete isolation. Unlike traditional continual learning methods that require data from previous domains (rehearsal) or complex architectural routing, a GeoLayer is optimized for domain $\mathcal{D}_i$ using only the COA loss (Eq. \ref{Eq:convex_ortho_loss}) and the target domain's data. Because the "Social Contract" is enforced locally through the orthogonality constraint, the layer is inherently "pre-compensated" for future stacking. This enables a modular workflow where new domains can be learned independently and integrated into the GeoStack asynchronously without retraining existing layers.
\section{Experimental Methodology}
\label{sec:experiments}
We evaluate the efficiency of GeoStack on two Vision problems: Multi-Domain Adaptation (MDA) and Class-Incremental Learning (CIL). The objective is to quantify the performance of GeoStack across disparate knowledge domains and its ability to handle catastrophic forgetting.

\textbf{Implementation:}
All experiments were conducted on a NVIDIA GeForce RTX 2080 Ti GPU. We use OpenCLIP's \textbf{ViT-B/16} as the backbone encoder, keeping all weights frozen while learning task-specific GeoLayers. Since each GeoLayer is constrained to an upper-triangular matrix $W \in \mathbb{R}^{d \times d}$, this reduces the learnable parameters by approximately 50\% ($\approx 1.3 \times 10^5$ parameters) compared to a full $d \times d$ transformation. Each GeoLayer is trained in isolation, independent of other domains, ensuring constant training complexity regardless of the total number of domains. We utilize the AdamW optimizer with a learning rate of $1 \times 10^{-4}$ and a batch size of 32, and train for 30–50 epochs. For the COA loss, we fix $\lambda = 0.95$ for all datasets and increase it to 0.99 for domain-specific datasets. 
% All experiments were conducted on an NVIDIA GeForce RTX 2080 Ti. We use OpenCLIP's ViT-B/16 as our foundational backbone, keeping all weights frozen while learning task-specific $512 \times 512$ GeoLayer matrices. A defining characteristic of our approach is that each GeoLayer is trained in isolation, independent of other domains; this allows for a flexible training strategy with minimal complexity.
% We utilize the AdamW optimizer with a learning rate of $1 \times 10^{-4}$ and a batch size of 32, train for 30-50 epochs depending on the complexity and the size of the datasets. For the COA loss (Eq. \ref{Eq:convex_ortho_loss}), we fix $\lambda = 0.95$ for all datasets and increase it to 0.99 for domain-specific datasets such as dtd. 

\subsection{Multi-Domain Adaptation (MDA)}
In this problem setting, we aim to adapt a foundation model to multiple target domains $\{\mathcal{D}_1, \mathcal{D}_2, \dots, \mathcal{D}_n\}$ simultaneously. Traditional MDA often requires joint training on data from all domains. The modular nature of GeoStack makes it a great candidate for MDA, where GeoLayers can be trained on individual domains separately and then stacked on each other to create a unified multi-domain model. This approach enables the model to perform well across all target domains without requiring simultaneous access to the data or complex joint optimization.

% We train  GeoLayers $W_i$ for each specialized domain independently using a few-shot protocol ($16$-shot). The objective is to evaluate if these independent units can be composed into a single, unified projection head $W_{total} = W_n \dots W_1 \cdot W_{base}$ that performs well across all constituent domains.

\textbf{Dataset Categorization:} To evaluate GeoStack, we curate a suite of datasets representing diverse semantic complexities. The datasets are categorized as: \textbf{General Objects} consisting of ImageNet-1K ($i$)~\citep{imagenet} and Caltech-101 ($c$)~\citep{caltech101}, \textbf{Fine-Grained Objects} consisting of Flowers-102 ($f$)~\citep{flowers102} ($f$) and Food-101 ($fo$)~\citep{food101}, and \textbf{Domain-Specific Images} consisting of EuroSAT ($e$)~\citep{eurosat}, DTD ($d$)~\citep{dtd}. This selection allows us to quantify the model's capacity to preserve foundational general-object knowledge while simultaneously adapting to fine-grained domains and specialized distributions.

% \textbf{Dataset Categorization:}
% To evaluate the performance of GeoStack, we curate a select suite of datasets representing various levels of semantic and geometric complexity. The datasets are categorized as: \textbf{General Object domain}, consisting of ImageNet-1k~\cite{imagenet} ($i$) and Caltech-101~\cite{caltech101} ($c$); \textbf{Fine-Grained domain} consisting of Flowers-102~\cite{flowers102} ($f$) and Food-101~\cite{food101} ($fo$); and \textbf{Domain-Specific (Distribution Shift)} consisting of EuroSAT~\cite{eurosat} ($e$) and DTD~\cite{dtd} ($d$). This allow us to evaluate the models capability to preserve the foundational knowldeg of the geenral objects while being simultaneously adapted to finegrained or domain specific tasks. 

% \begin{itemize}
%     \item \textbf{General Object Recognition}: Includes ImageNet-1k~\cite{imagenet} ($i$) and Caltech-101~\cite{caltech101} ($c$) and contains general objects. This serves as the foundational domain knowledge that must be preserved under knowledge composition.

%     \item \textbf{Fine-Grained Classification}: Includes Flowers-102~\cite{flowers102} ($f$) and Food-101~\cite{food101} ($fo$). These contain narrow semantic categories that must be integrated without disrupting the foundational knowledge.

%     \item \textbf{Domain-Specific (Distribution Shift)}: Includes EuroSAT~\cite{eurosat} ($e$) and DTD~\cite{dtd} ($d$). These samples depart significantly from the natural image distribution of CLIP's pre-training, and act as a geometric stability test for the GeoStack.
% \end{itemize}

\subsubsection{Multi-Domain Knowledge Composition with GeoStack}
We adopt a two-stage process for knowledge integration, inspired by AdapterFusion~\citep{pfeiffer2021adapterfusion}:

\textbf{Stage 1 - Knowledge Extraction:} For each domain $\mathcal{D}_k$, a dedicated GeoLayer is trained in isolation using the COA loss under a 16-shot (16 samples per class) protocol. This stage extracts domain-specific expertise while ensuring it is composable.

\textbf{Stage 2 - Knowledge Composition:} To create a unified multi-expert model, we compose the independent GeoLayers into a single \textbf{GeoStack}. For example, a stack sequence denoted as $i \to c \to fo \to e$ computes the final transformation as: $W_{g} = W_i \cdot W_c \cdot W_{fo} \cdot W_e$. Here, the arrows ($\to$) denote the stacking order, where the initial GeoLayer for ($i$) forms the base and subsequent layers are appended to the transformation chain. The geometric constraints enforced during Stage 1 ensure that the resulting product $W_{g}$ remains stable for all domains.

\subsubsection{QuadStack: Results and Discussion} 
While GeoStack can be composed at any arbitrary depth, our analysis focuses on a \textbf{Quad-Stack} (Depth-4) configuration. A dual or triple stack may often fail to reveal the potential for long-term instability. By evaluating a Depth-4 composition, we demonstrate GeoStack’s capability to maintain cross-domain performance while effectively thwarting catastrophic forgetting.

To evaluate the stability of GeoStack, we define three stacks of increasing domain complexity: (1) the \textbf{Easy Stack} ($i \to c \to fo \to e$), following a coarse-to-fine semantic transition; (2) the \textbf{Moderate Stack} ($i \to fo \to e \to d$), representing a progression to increasing complexity; and (3) the \textbf{Hard Stack} ($i \to e \to d \to f$) representing a departure from general image domain to domain specific visual content. 
\begin{table*}[th]
\centering
\small
\begin{tabular}{@{}llccccc@{}}
\toprule
\textbf{Stack} & \textbf{Dataset} & \textbf{ZS} & \textbf{TA} & \textbf{TA} & \textbf{BiCLIP} & \textbf{GeoStack} \\
 &  & & \textbf{(BiCLIP)} & \textbf{(Geo)} & [OE] & [OE] \\ \midrule

\multirow{5}{*}{\shortstack[l]{\textbf{Easy Stack} \\ $i \to c \to fo \to e$}} 
    & ImageNet ($i$) & 66.6\% & 60.2\% & 66.2\% & 67.1\% [0.022] & \textbf{69.3\% [0.010]} \\
    & Caltech-101 ($c$) & 90.0\% & 87.6\% & 90.0\% & 92.9\% [0.022] & \textbf{93.1\% [0.010]} \\
    & Food-101 ($fo$) & 88.7\% & 85.4\% & 88.0\% & 88.2\% [0.022] & \textbf{89.5\% [0.010]} \\
    & EuroSAT ($e$) & 47.5\% & 31.5\% & 37.8\% & \textbf{84.9\% [0.022]} & 84.1\% [0.010] \\ \cmidrule(l){2-7} 
    & \textbf{Average} & 73.2\% & 66.2\% & 70.5\% & 83.3\% [0.022] & \textbf{84.0\% [0.010]} \\ \midrule

\multirow{5}{*}{\shortstack[l]{\textbf{Moderate Stack} \\ $i \to fo \to e \to d$}} 
    & ImageNet ($i$) & 66.6\% & 62.0\% & 67.9\% & 57.2\% [0.052] & \textbf{65.4\% [0.012]} \\
    & Food-101 ($fo$) & 88.7\% & 85.3\% & 88.5\% & 82.6\% [0.052] & \textbf{87.1\% [0.012]} \\
    & EuroSAT ($e$) & 47.5\% & 83.8\% & 82.6\% & 83.2\% [0.052] & \textbf{83.3\% [0.012]} \\
    & DTD ($d$) & 42.8\% & 39.3\% & 42.4\% & \textbf{69.7\% [0.052]} & 66.7\% [0.012] \\ \cmidrule(l){2-7} 
    & \textbf{Average} & 61.4\% & 67.6\% & 70.4\% & 73.2\% [0.052] & \textbf{75.6\% [0.012]} \\ \midrule

\multirow{5}{*}{\shortstack[l]{\textbf{Hard Stack} \\ $i \to e \to d \to f$}} 
    & ImageNet ($i$) & 66.6\% & 49.0\% & 62.1\% & 52.6\% [0.070] & \textbf{62.8\% [0.013]} \\
    & EuroSAT ($e$) & 47.5\% & 82.0\% & 78.3\% & 81.7\% [0.070] & \textbf{82.8\% [0.013]} \\
    & DTD ($d$) & 42.8\% & 67.7\% & 60.0\% & \textbf{69.5\% [0.070]} & 66.1\% [0.013] \\
    & Flowers-102 ($f$) & 71.0\% & 52.3\% & 60.9\% & \textbf{86.8\% [0.070]} & 85.8\% [0.013] \\ \cmidrule(l){2-7} 
    & \textbf{Average} & 57.0\% & 62.8\% & 65.3\% & 72.6\% [0.070] & \textbf{74.4\% [0.013]} \\ \bottomrule
\end{tabular}
\caption{\textbf{Quad-Stack Multi-Domain Adaptation results.} We evaluate stackability across three tiers of geometric difficulty, comparing Task Arithmetic (TA) against  GeoStack.\label{tab:mda_results}}
\end{table*} 

We compare GeoStack against Task Arithmetic~\citep{ilharco2023editingmodelstaskarithmetic} (TA) by linearly summing the learned perturbations $\Delta_i$. TA treats knowledge composition as an additive operation ($\mathbf{I} + \alpha \sum_i\Delta_i$). We empirically observe that while $\alpha$ is often tuned in TA to balance task performance and interference, setting $\alpha$ to match the orthogonality error (OE) of GeoStack results in a degradation of performance. Consequently, we set $\alpha=1$ for our primary baseline comparisons. We compare classification accuracy on five configurations: \textbf{1. ZS:} The vanilla zero-shot CLIP model without any adapters, \textbf{2. TA (BiCLIP):} Linear task arithmetic using bilinear adapters. \textbf{3. TA (Geo):} Linear task arithmetic using GeoLayers (trained with COA loss). \textbf{4. BiCLIP:} A naive stacking baseline where each domain expert is trained as a standard bilinear adapter $W_i$ without the orthogonality constraint ($\lambda = 0$), and \textbf{5. GeoStack [OE] (Proposed):} Where GeoLayers are trained with the COA loss and then stacked. 

The results, summarized in Table \ref{tab:mda_results}, demonstrate an intuitive correlation between the geometric complexity of target domains, the accumulation of Orthogonality Error (OE), and the preservation of foundational knowledge. Notably, Task Arithmetic (TA) with the constrained GeoLayers yields significant gains over unconstrained BiCLIP, ImageNet accuracy improves from $49.0\%$ to $62.1\%$ in the Hard Stack. In BiCLIP, the introduction of out-of-distribution domains like EuroSAT ($e$) and DTD ($d$) results in a collapse of the foundational knowledge. In the Hard Stack, ImageNet accuracy plummets from $66.6\%$ to $52.6\%$ as the cumulative OE increases to $0.070$. This confirms that without geometric constraints, subsequent experts distort the existing knowledge of previous domains.  Conversely, GeoStack, attributing to the COA loss, maintains ImageNet accuracy at $62.8\%$ with an OE of $0.013$. GeoStack consistently provides superior \textbf{Average} classification accuracy by maintaining the foundational knowledge, thus validating that geometric constraints are required for stable, modular multi-domain composition.

\textit{Additional results for dual, triple, and hexa-stack are provided in Appendix~\ref{Appendix:B}.}
% While the unconstrained baseline occasionally achieves marginal gains ($\sim1\%$) on the final task, it does so by sacrificing the utility of the entire stack. 

\subsection{Class-Incremental Learning (CIL)}
In this setting, the objective is to learn a sequence of novel classes without sacrificing the ability to recognize previously learned ones. Catastrophic forgetting remains a primary challenge in CIL, resulting in a model that performs poorly on older tasks.

GeoStack offers a unique architectural solution for knowledge composition in CIL. GeoLayer can be trained for each incremental batch of classes and composed into a stack, while keeping the backbone parameters frozen. This makes CIL an ideal problem for quantifying catastrophic forgetting. GeoLayers can be considered as non-disruptive if the composed margin remains wide enough to incorporate new class features while maintaining the performance on the original classes.

\subsubsection{Knowledge composition for Class-Incremental Learning}
We conduct experiments on CIFAR-100~\citep{krizhevsky2009learning}, consisting of 100 categories of general objects. In this experiment, we define $n$ disjoint tasks $\{\mathcal{T}_0, \mathcal{T}_1, \dots, \mathcal{T}_{n-1}\}$, where each task introduces a new set of classes. We set $n=4$ and partition the dataset into four tasks consisting of 25 disjoint classes. For each task $\mathcal{T}_k$, we train a GeoLayer $W_k$ using a 16-shot protocol independently. The cumulative knowledge of the model after $n$ tasks is represented by the composite operator: $W_{g} = \prod_{k=0}^{n-1} W_k$.

We quantify the model's capability for Incremental Learning and Knowledge Retention. In the former, we measure the model’s ability to recognize all classes seen so far (Global Accuracy); in the latter, we measure the model's ability to retain knowledge from the initial classes as newer tasks are added to the stack (Task-0 Retention).

\begin{table}[ht]
\centering
\small
\setlength{\tabcolsep}{5pt}
\begin{minipage}[t]{0.48\textwidth}
\begin{tabular}{@{}lccccc@{}}
\toprule
\textbf{Task} & \textbf{Classes} & \textbf{ZS} & \textbf{BiCLIP} & \textbf{TA} & \textbf{GeoStack} \\ \midrule
T0 & 25  & 71.88 & \textbf{86.20} & 77.92          & 77.92 \\
T1 & 50  & 68.44 & \textbf{74.60} & 69.96          & 74.18 \\
T2 & 75  & 66.89 & 63.99          & 65.76          & \textbf{70.47} \\
T3 & 100 & 68.11 & 60.08          & 61.79          & \textbf{69.47} \\ \bottomrule
& & & & & \\
\end{tabular}
\end{minipage}
\hfill
\begin{minipage}[t]{0.48\textwidth}
\centering
\begin{tabular}{@{}lccc@{}}
\toprule
\textbf{Evaluation} & \textbf{BiCLIP} & \textbf{TA} & \textbf{GeoStack} \\ \midrule
After T0 & \textbf{86.20} & 77.92 & 77.92 \\
After T1 & 81.88          & 79.48 & \textbf{77.20} \\
After T2 & 76.60          & \textbf{77.28} & 76.92 \\
After T3 & 72.04          & 74.00 & \textbf{75.80} \\ \midrule
\textbf{Decay ($\Delta$)} & \color{red}{-14.16} & -3.92 & \color{blue}{\textbf{-2.12}} \\ \bottomrule
\end{tabular}
\end{minipage}
\vspace{1em}
\caption{\textbf{CIFAR-100 Incremental Learning Results.} \textit{Left:} Global accuracy on all seen classes. \textit{Right:} Retention of Task-0 (first 25 classes). Zero-Shot (ZS) baseline for Task-0 is 71.88\%. \label{tab:cil_combined}}
\end{table}\vspace{-2em}

% \begin{table}[ht]
% \centering
% \small
% \caption{\textbf{Global Incremental Accuracy (CIFAR-100).} Evaluation on the cumulative set of all classes seen. By Task 2, BiCLIP falls below the Zero-Shot baseline.}
% \label{tab:cil_global}
% \begin{tabular}{@{}lcccc@{}}
% \toprule
% \textbf{Task} & \textbf{Classes} & \textbf{Zero-Shot} & \textbf{BiCLIP} & \textbf{GeoStack} \\ \midrule
% Task 0 & 25  & 71.88\% & \textbf{86.20\%} & 77.92\% \\
% Task 1 & 50  & 68.44\% & \textbf{74.60\%} & 74.18\% \\
% Task 2 & 75  & 66.89\% & 63.99\% & \textbf{70.47\%} \\
% Task 3 & 100 & 68.11\% & 60.08\% & \textbf{69.47\%} \\ \bottomrule
% \end{tabular}
% \end{table}

% \begin{table}[ht]
% \centering
% \small
% \caption{\textbf{Task-0 Retention (CIFAR-100).} Accuracy on the initial 25 classes as the stack grows. GeoStack prevents the "catastrophic forgetting" seen in the baseline.}
% \label{tab:cil_retention}
% \begin{tabular}{@{}lccc@{}}
% \toprule
% \textbf{Evaluation Stage} & \textbf{Zero-Shot} & \textbf{BiCLIP} & \textbf{GeoStack} \\ \midrule
% After Task 0 & 71.88\% & \textbf{86.20\%} & 77.92\% \\
% After Task 1 & 71.88\% & 81.88\% & \textbf{77.20\%} \\
% After Task 2 & 71.88\% & 76.60\% & \textbf{76.92\%} \\
% After Task 3 & 71.88\% & 72.04\% & \textbf{75.80\%} \\ \midrule
% \textbf{Total Decay ($\Delta$)} & -- & \color{red}{-14.16\%} & \color{blue}{-2.12\%} \\ \bottomrule
% \end{tabular}
% \end{table}

\subsubsection{QuadStack: Results and Discussion}
The results in Table~\ref{tab:cil_combined} (left) demonstrate that GeoStack's ability to facilitate knowledge composition significantly minimizes catastrophic forgetting. While the BiCLIP baseline initially achieves a high 86.20\% accuracy on Task 0, this performance degrades rapidly as the stack depth increases. In the incremental learning setting, BiCLIP's accuracy drops by approximately 8.7\% per additional task, eventually falling to 60.08\%—which is 8\% below the Zero-Shot CLIP baseline. Similarly, TA (using GeoLayers) falls below Zero-Shot by T2. This indicates that without geometric constraints, the models are susceptible to catastrophic forgetting.

In contrast, GeoStack maintains an accuracy of 69.47\% after composing knowledge from four independently trained tasks, outperforming the baseline by 9.39\%. The Retention experiment in Tables \ref{tab:cil_combined} (right) further highlights GeoStack's resistance to catastrophic forgetting: while BiCLIP loses 14.16\% of its original knowledge, TA loses 3.92\%, while GeoStack’s geometric constraint limits it to 2.12\% (77.92\% to 75.80\%). These findings are particularly encouraging, as successful knowledge composition from independently trained models remains a complex problem in current literature. We attribute this success to the inherent robustness of CLIP's feature space and the effectiveness of the geometric constraints in preserving the previous knowledge successfully.

\subsection{Analysis of GeoStack Properties}
In this section, we validate the properties of GeoStack by testing its Abelian nature and performing a deep-stacking stress test to examine the margin erosion.

\textbf{1. The Abelian Property:}
\label{sec:abelian_property}
\begin{figure}[th]
    \centering
    \begin{minipage}{0.54\textwidth}
        \centering
        \includegraphics[width=.7\textwidth]{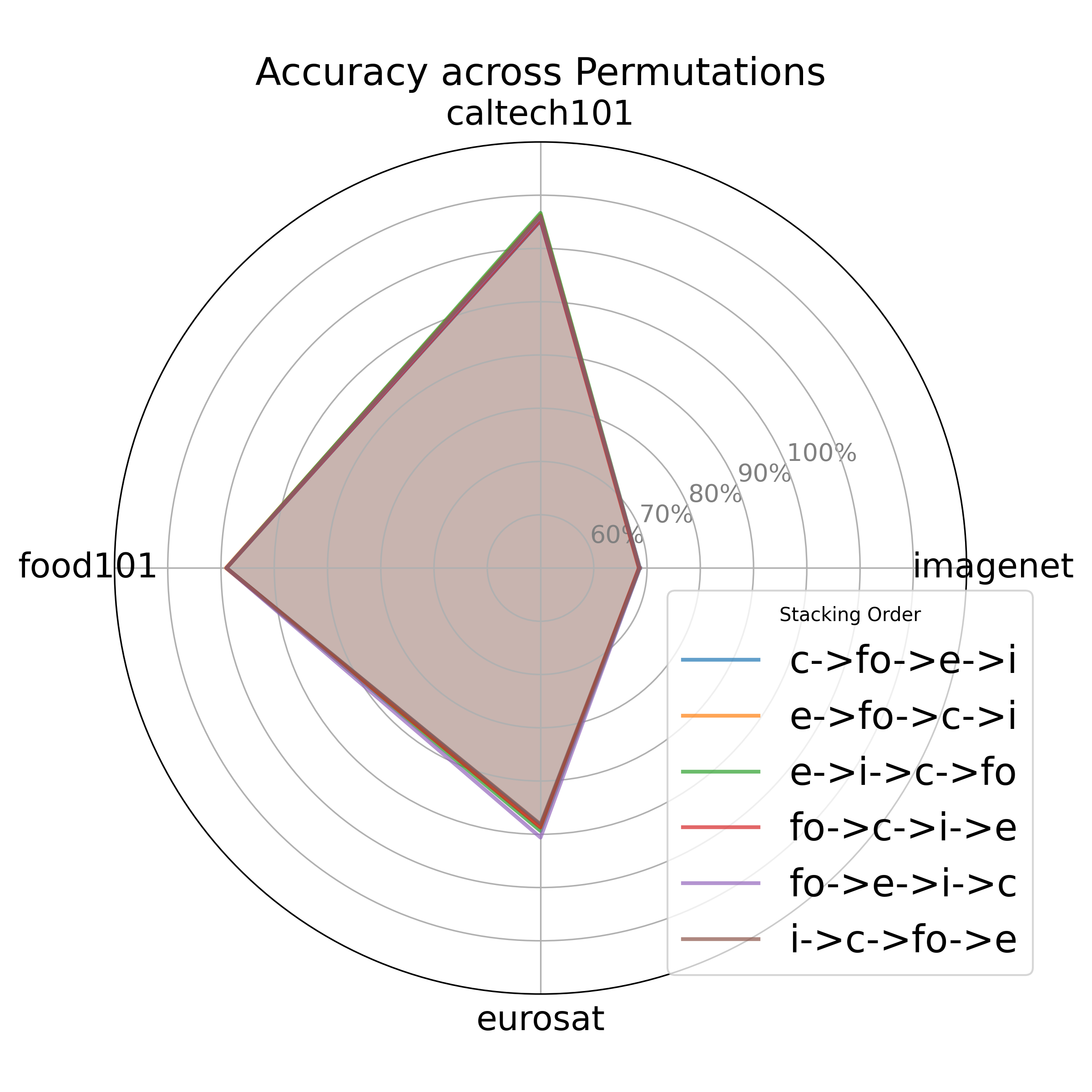}
        \captionof{figure}{Spider plot showing accuracy invariance across four stacking permutations.}
        \label{fig:abelian_combined}
    \end{minipage}
    \hfill
    \begin{minipage}{0.45\textwidth}
        \centering
        \small
        \begin{tabular}{lccc}
            \toprule
            \textbf{Dataset} & \textbf{Mean (\%)} & \textbf{$\sigma$} & \textbf{Range} \\
            \midrule
            ImageNet & 69.29 & 0.04 & 0.09 \\
            Caltech101 & 92.97 & 0.27 & 0.74 \\
            Food101 & 89.49 & 0.02 & 0.07 \\
            EuroSAT & 84.49 & 0.42 & 1.25 \\
            \bottomrule            
        \end{tabular}        
        \captionof{table}{Accuracy statistics over permutations.\label{tab:abelian_variance}}
    \end{minipage}    
\end{figure}
The Abelian property (commutativity) is a highly desirable characteristic for modular knowledge composition. It allows the GeoLayers to be stacked in any arbitrary order without the need for a combinatorial optimization. To validate this property, we evaluate a QuadStack configuration consisting of four domain experts from MDA: $\mathcal{D}_{ImgNet} (i)$, $\mathcal{D}_{Caltech} (c)$, $\mathcal{D}_{Food} (fo)$, and $\mathcal{D}_{Euro} (e)$. We measure the performance across multiple stacking permutations to observe sensitivity to ordering. 

As illustrated in the spider plot (Fig.~\ref{fig:abelian_combined}), the performance for each domain remains consistent regardless of the sequence. For instance, the accuracy on EuroSAT remains stable at $84.49 \pm 0.42\%$, even when its position in the stack is shifted from the base to the final layer. As shown in Table~\ref{tab:abelian_variance}, the negligible variance across all tested permutations confirms the Quasi-Abelian nature of GeoStack under the geometric constraints.

\textbf{2. Margin Erosion and Deep Stacking:} 
\label{sec:margin_erosion}
To quantify GeoStack's behavior under deep-stacking conditions, we conduct a 10-task class-incremental learning experiment on the CIFAR-100 dataset. The dataset is partitioned into 10 sequential tasks, each containing 10 disjoint classes. We measure the retention of Task-0 accuracy and the ImageNet foundational knowledge to quantify catastrophic forgetting as the stacking depth increases from $n=1$ to $n=10$.
% \begin{figure}[th]
%   \begin{minipage}{0.54\textwidth}
%     \centering\includegraphics[width=.65\linewidth]{images/depth_sustainability_plot.png}    
%   \end{minipage}
%   \begin{minipage}{0.44\textwidth}    
%     \caption{Comparison of Task-0 accuracy and Imagenet foundational knowledge retention across 10 incremental learning tasks. 
%     % While BiCLIP (red) suffers from an exponential decay with each new task, GeoStack (blue) maintains a graceful degradation, sustaining significantly higher robustness to catestrophic forgetting.
%     \label{fig:depth_decay}}
%   \end{minipage}
% \end{figure}

% \begin{figure}[th]
%     \centering
%     \includegraphics[width=0.48\textwidth]{images/depth_sustainability_plot.png}
%     \caption{\textbf{Manifold Sustainability.} Comparison of Task-0 accuracy retention across 10 sequential tasks. While BiCLIP (red) suffers from exponential decay as the geometric manifold collapses, GeoStack (blue) maintains a graceful degradation, sustaining significantly higher retention through its orthogonality constraint.}
%     \label{fig:depth_decay}
% \end{figure}

\begin{figure}[th]
  \centering  
  \begin{minipage}{0.48\textwidth}
    \centering
    \includegraphics[width=.9\linewidth]{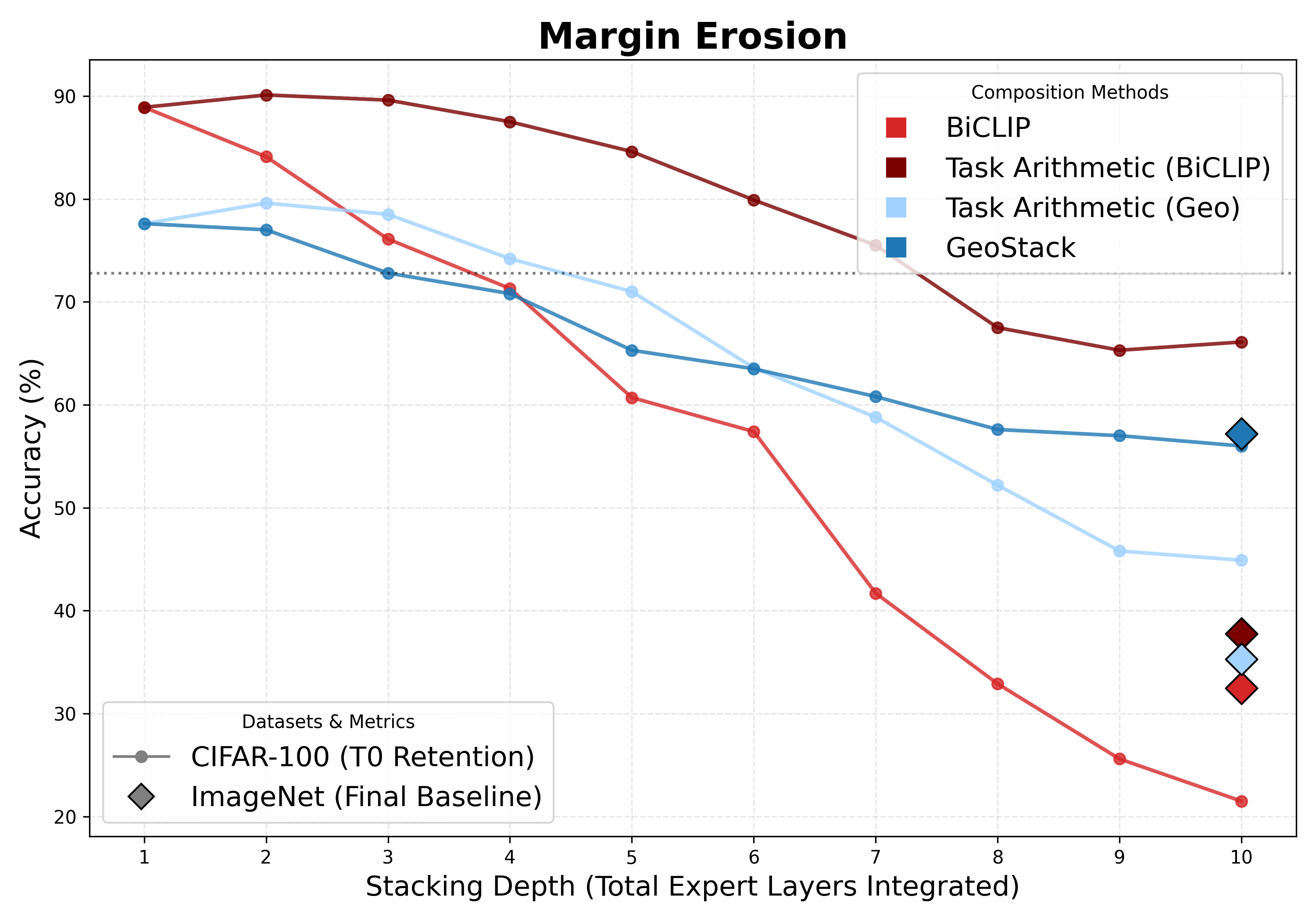}
    \caption{Comparison of Task-0 accuracy and ImageNet knowledge retention across 10 incremental tasks.}
    \label{fig:depth_decay}
  \end{minipage}
  \hfill % Adds spacing between the two figures
  % --- Second Figure (Right) ---
  \begin{minipage}{0.48\textwidth}
    \centering
    \includegraphics[width=.9\linewidth]{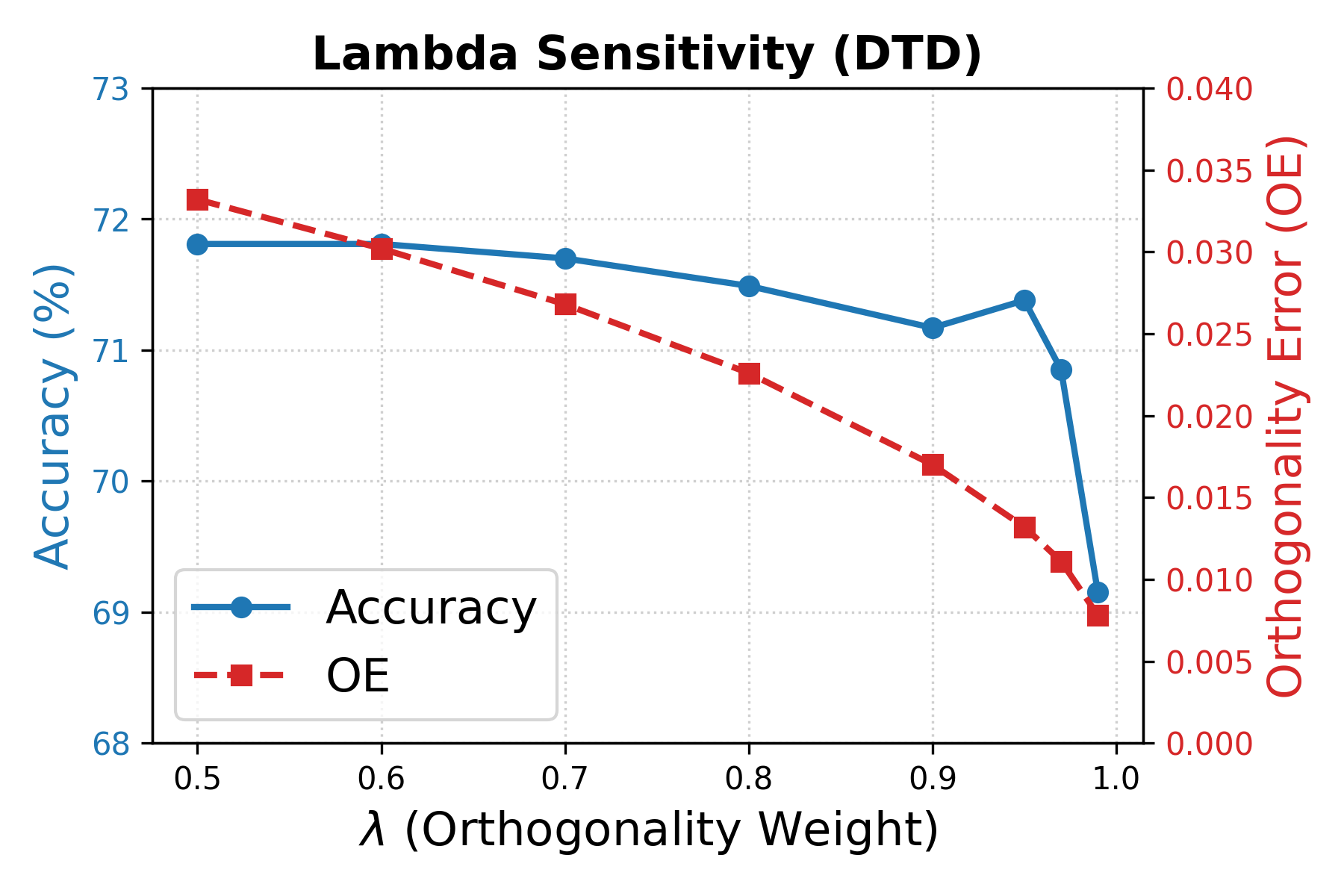} % Replace with your second image path
    \caption{Sensitivity analysis of the geometric constraint weight $\lambda$ on the DTD dataset.}
    \label{fig:lambda_sensitivity}
  \end{minipage}
\end{figure}

As illustrated in Figure~\ref{fig:depth_decay}, GeoStack exhibits robust resistance to catastrophic forgetting; it retains an accuracy of 56.00\% on the initial task, significantly outperforming the BiCLIP baseline, which drops to 21.50\%. The results indicate that while BiCLIP initially achieves higher peak performance on the first task ($88.9\%$), it quickly leads to an exponential decay with stacking depth. In contrast, GeoStack exhibits a steady and linear degradation in performance. Most notably, GeoStack preserves the model's foundational integrity with a final ImageNet accuracy of 57.2\%, representing a 19.4\% margin over Task Arithmetic with BiCLIP. This confirms that GeoStack preserves foundational knowledge under deep stacking.

% \subsubsection{Failure Case?}
\textbf{3. Sensitivity Analysis of $\lambda$:} 
We perform a sensitivity analysis on $\lambda$, varying it in the range $0.5$ to $0.99$. Results shown in Fig~\ref{fig:lambda_sensitivity} demonstrate that increasing $\lambda$ yields an exponential reduction in Orthogonality Error—decreasing from $0.0332$ to $0.0078$—with a marginal $2.6\%$ impact on accuracy. This confirms that a high $\lambda$ successfully enables stackability, which is crucial for maintaining foundational integrity.

\textit{Further analysis on the Stacking metric is provided in Appendix~\ref{Appendix:C}.}
\section{Conclusion}
In this work, we presented GeoStack, a modular framework that solves the trade-off between knowledge accumulation and catastrophic forgetting in VLM adapters. By enforcing geometric constraints through a novel Convex Orthogonality Alignment (COA) loss, we developed GeoLayers that are stackable for knowledge composition while preserving the foundational CLIP latent space. We demonstrated that an arbitrary number of domain experts can be folded into a single $O(1)$ projection matrix. Experiments on Multi-Domain Adaptation and Class-Incremental Learning reveal that GeoStack matches the performance of domain-specific models while significantly mitigating catastrophic forgetting.

% \section*{References}

{
    \small
    \bibliographystyle{abbrvnat}
    \bibliography{main}
    
}

% {
% \small

% [1] Alexander, J.A.\ \& Mozer, M.C.\ (1995) Template-based algorithms for
% connectionist rule extraction. In G.\ Tesauro, D.S.\ Touretzky and T.K.\ Leen
% (eds.), {\it Advances in Neural Information Processing Systems 7},
% pp.\ 609--616. Cambridge, MA: MIT Press.

% [2] Bower, J.M.\ \& Beeman, D.\ (1995) {\it The Book of GENESIS: Exploring
%   Realistic Neural Models with the GEneral NEural SImulation System.}  New York:
% TELOS/Springer--Verlag.

% [3] Hasselmo, M.E., Schnell, E.\ \& Barkai, E.\ (1995) Dynamics of learning and
% recall at excitatory recurrent synapses and cholinergic modulation in rat
% hippocampal region CA3. {\it Journal of Neuroscience} {\bf 15}(7):5249-5262.
% }

\newpage
\appendix
\section*{Appendix A: Theoretical Derivations.}
\renewcommand{\thesubsection}{A.\arabic{subsection}}
\setcounter{subsection}{0}
\subsection{Relation between Orthogonality Error and Spectral Norm}
\label{Appendix:A}
In GeoStack, under the structural constraint, GeoLayers are initialized as Identity matrices and constrained to be upper-triangular. GeoLayer parameters can be represented as $W = \mathbf{I} + \Delta$, where $\mathbf{I}$ is the identity matrix and $\Delta$ is an upper-triangular matrix representing a learned perturbation. 

From Sec.~\ref{sec:perturbation_minimization}, we know that the stability condition will hold if the spectral norm of the perturbation $\|\Delta\|_2$ is small. 

We define Orthogonality Error (OE) as the Frobenius norm of deviation from the Identity matrix. This is given by:
$$\mathcal{L}_{ortho} = \|W^\top W - \mathbf{I}\|_F^2$$

Substituting $W = I + \Delta$:
\begin{equation}
    \begin{split}
        \|W^\top W - \mathbf{I}\|^2_F &= \|(\mathbf{I} + \Delta^\top)(\mathbf{I} + \Delta) - \mathbf{I}\|^2_F \\     
        & = \|\mathbf{I} + \Delta + \Delta^\top + \Delta^\top\Delta - \mathbf{I}\|^2_F \\
        & = \|\Delta + \Delta^\top + \Delta^\top\Delta\|^2_F
    \end{split}
\end{equation}

Under the assumption that $W$ is initialized to the identity and the perturbations are small, the second-order term $\Delta^\top\Delta$ is negligible. Thus, we approximate:
$$\mathcal{L}_{ortho} \approx \|\Delta + \Delta^\top\|^2_F$$

Since $(\Delta + \Delta^\top)$ is symmetric, its entries are $d_{ii} + d_{ii}$ on the diagonal and $u_{ij}$ or $u_{ji}$ on the off-diagonals. One can see that:
$$\|\Delta + \Delta^\top\|_F^2 = 2 \|\Delta\|_F^2 + 2 \sum_{i} d_{ii}^2$$

Thus, we establish a lower bound relative to the perturbation:
$$\|\Delta + \Delta^\top\|^2_F \geq 2\|\Delta\|^2_F$$
$$ \implies \mathcal{L}_{ortho} \gtrsim 2\|\Delta\|^2_F $$

Finally, since the Frobenius norm upper bounds the spectral norm ($\|\Delta\|_2 \leq \|\Delta\|_F$), we have:
$$2\|\Delta\|^2_2 \leq 2\|\Delta\|^2_F \lesssim \mathcal{L}_{ortho}$$

$$ \implies \|\Delta\|_2 \lesssim \sqrt{\frac{\mathcal{L}_{ortho}}{2}} $$

Consequently, by minimizing the Orthogonality Error $\mathcal{L}_{ortho}$, the optimization process suppresses the growth of the spectral norm of the perturbation $\|\Delta\|_2$. Ultimately, this fulfills the stability condition required for GeoStack, allowing multiple domain experts to be integrated without the destructive interference from one another.

\section*{Appendix B: Scaling Behavior.} 
\renewcommand{\thesubsection}{B.\arabic{subsection}}
\setcounter{subsection}{0}
\subsection{Dual Stack Analysis}
\label{Appendix:B}
In this section, we present the performance of GeoStack for MDA using dual-stack compositions. We assume ImageNet as the foundational expertise and quantify the performance of GeoStack under an additional cross-domain composition. We compare the standard BiCLIP adapter (baseline) against the GeoStack framework. Stability is measured via the Orthogonality Error (OE) and the retention of foundational knowledge on ImageNet.

\begin{table}[ht]
\centering
\caption{Dual-Stack Performance Comparison. Brackets $[ \cdot ]$ denote the Orthogonality Error (OE). Zero-shot values represent the performance of the frozen CLIP backbone.}
\label{tab:appendix_dual_stack}
\small
\begin{tabular}{lcccc}
\toprule
\textbf{Stack Sequence} & \textbf{Dataset} & \textbf{Zero-Shot} & \textbf{BiCLIP (Baseline)} & \textbf{GeoStack (Ours)} \\ \midrule

\multirow{3}{*}{\shortstack[l]{\textbf{Sequence A} \\ $i \to c$}} 
    & ImageNet ($i$) & 66.6\% & 69.1\% [0.011] & \textbf{69.6\% [0.006]} \\
    & Caltech-101 ($c$) & 90.0\% & 93.5\% [0.011] & \textbf{93.9\% [0.006]} \\ \cmidrule(l){2-5}
    & \textbf{Average} & 78.3\% & 81.3\% & \textbf{81.8\%} \\ \midrule

\multirow{3}{*}{\shortstack[l]{\textbf{Sequence B} \\ $i \to fo$}} 
    & ImageNet ($i$) & 66.6\% & 69.6\% [0.011] & \textbf{70.2\% [0.007]} \\
    & Food-101 ($fo$) & 88.7\% & 89.6\% [0.011] & \textbf{90.0\% [0.007]} \\ \cmidrule(l){2-5}
    & \textbf{Average} & 77.7\% & 79.6\% & \textbf{80.1\%} \\ \midrule

\multirow{3}{*}{\shortstack[l]{\textbf{Sequence C} \\ $i \to e$}} 
    & ImageNet ($i$) & 66.6\% & 67.6\% [0.023] & \textbf{69.6\% [0.007]} \\
    & EuroSAT ($e$) & 47.5\% & \textbf{85.7\% [0.023]} & 84.7\% [0.007] \\ \cmidrule(l){2-5}
    & \textbf{Average} & 57.1\% & 76.7\% & \textbf{77.2\%} \\ \midrule

\multirow{3}{*}{\shortstack[l]{\textbf{Sequence D} \\ $i \to d$}} 
    & ImageNet ($i$) & 66.6\% & 59.5\% [0.050] & \textbf{66.3\% [0.009]} \\
    & DTD ($d$) & 42.8\% & \textbf{69.7\% [0.050]} & 68.4\% [0.009] \\ \cmidrule(l){2-5}
    & \textbf{Average} & 54.7\% & 64.6\% & \textbf{67.4\%} \\ \midrule

\multirow{3}{*}{\shortstack[l]{\textbf{Sequence E} \\ $i \to f$}} 
    & ImageNet ($i$) & 66.6\% & 65.8\% [0.026] & \textbf{67.9\% [0.010]} \\
    & Flowers ($f$) & 71.0\% & \textbf{94.2\% [0.026]} & 92.7\% [0.010] \\ \cmidrule(l){2-5}
    & \textbf{Average} & 68.8\% & 80.0\% & \textbf{80.3\%} \\ \bottomrule
\end{tabular}
\end{table}

The dual-stack experiments (Table~\ref{tab:appendix_dual_stack}) reveal a clear correlation between Orthogonality Error (OE) and foundational knowledge retention. Across all five sequences, GeoStack consistently maintains higher ImageNet accuracy compared to the BiCLIP baseline. This is most evident in Sequence D (DTD), where the baseline's OE explodes to $0.050$, resulting in a catastrophic $7.1\%$ drop in ImageNet performance compared to the zero-shot baseline. In contrast, GeoStack restricts the OE to $0.009$, preserving the foundational knowledge within $0.4\%$ of its original state.

\subsection{Triple Stack Analysis}
In this section, we present the performance of GeoStack for MDA using three triple-stack compositions simialr to the quad-stack compositions: Easy ($i \to c \to fo$), Moderate ($i \to fo \to e$), and Hard ($i \to d \to f$).

\begin{table}[ht]
\centering
\caption{Triple-Stack Performance Comparison. The results highlight how Orthogonality Error (OE) accumulates across three-expert compositions.}
\label{tab:appendix_triple_stack}
\small
\begin{tabular}{lcccc}
\toprule
\textbf{Stack Sequence} & \textbf{Dataset} & \textbf{Zero-Shot} & \textbf{BiCLIP (Baseline)} & \textbf{GeoStack (Ours)} \\ \midrule

\multirow{4}{*}{\shortstack[l]{\textbf{Easy Stack} \\ $i \to c \to fo$}} 
    & ImageNet ($i$) & 66.6\% & 69.1\% [0.013] & \textbf{69.7\% [0.009]} \\
    & Caltech-101 ($c$) & 90.0\% & \textbf{93.4\% [0.013]} & 93.2\% [0.009] \\
    & Food-101 ($fo$) & 88.7\% & 89.4\% [0.013] & \textbf{89.7\% [0.009]} \\ \cmidrule(l){2-5}
    & \textbf{Average} & 81.8\% & 84.0\% & \textbf{84.2\%} \\ \midrule

\multirow{4}{*}{\shortstack[l]{\textbf{Moderate  Stack} \\ $i \to fo \to e$}} 
    & ImageNet ($i$) & 66.6\% & 67.7\% [0.022] & \textbf{69.7\% [0.009]} \\
    & Food-101 ($fo$) & 88.7\% & 88.3\% [0.022] & \textbf{89.6\% [0.009]} \\
    & EuroSAT ($e$) & 47.5\% & \textbf{84.5\% [0.022]} & 83.9\% [0.009] \\ \cmidrule(l){2-5}
    & \textbf{Average} & 67.6\% & 80.2\% & \textbf{81.1\%} \\ \midrule

\multirow{4}{*}{\shortstack[l]{\textbf{Hard  Stack} \\ $i \to d \to f$}} 
    & ImageNet ($i$) & 66.6\% & 55.2\% [0.062] & \textbf{63.6\% [0.013]} \\
    & DTD ($d$) & 42.8\% & \textbf{69.7\% [0.062]} & 67.2\% [0.013] \\
    & Flowers ($f$) & 71.0\% & \textbf{88.8\% [0.062]} & 87.2\% [0.013] \\ \cmidrule(l){2-5}
    & \textbf{Average} & 59.8\% & 71.2\% & \textbf{72.7\%} \\ \bottomrule
\end{tabular}
\end{table}

BiCLIP degrades under a triple stack composition and shows catastrophic forgetting. ImageNet accuracy drops to $55.2\%$ in the hard stack. GeoStack maintains an OE of only 0.013, keeping ImageNet accuracy at $63.6\%$. While there is a slight drop from zero-shot, it is $8.4\%$ higher than the baseline.

\subsection{Beyond Quad Stack}
To determine the upper limits of manifold stability, we evaluate on two additional datasets, StanfordCars~\citep{stanfordcars} (s), and Oxford-Pets~\citep{oxfordpets} (o). We evaluate the performance on Hexa-stack composition ($i \to o \to f \to s \to e \to d$). This sequence represents a significant accumulation of disparate domain knowledge, including fine-grained animals, flowers, vehicles, satellite imagery, and textures.

\begin{table}[ht]
\centering
\caption{Hexa-Stack Performance Comparison. OE denotes the final cumulative Orthogonality Error after all six experts are folded into the backbone.}
\label{tab:six_stack}
\small
\begin{tabular}{lcccc}
\toprule
\textbf{Dataset} & \textbf{Zero-Shot} & \textbf{BiCLIP (Baseline)} & \textbf{GeoStack (Ours)} \\ \midrule 
ImageNet ($i$)      & 66.6\% & 39.7\% & \textbf{62.2\%} \\
Oxford-Pet ($o$)    & 89.0\% & 72.7\% & \textbf{86.3\%} \\
Flowers-102 ($f$)   & 71.0\% & 71.0\% & \textbf{86.8\%} \\
Stanford-Cars ($s$) & 66.3\% & \textbf{65.2\%} & 60.2\% \\
EuroSAT ($e$)       & 47.5\% & 72.4\% & \textbf{81.7\%} \\
DTD ($d$)           & 42.8\% & 62.8\% & \textbf{63.2\%} \\ \midrule
\textbf{Average Acc.} & 63.9\% & 64.0\% & \textbf{73.4\%} \\
\textbf{Final OE $\downarrow$} & --- & 0.1359 & \textbf{0.0142} \\ 
\bottomrule
\end{tabular}
\end{table}

The results in Table~\ref{tab:six_stack} demonstrate the results for Hexa-stack composition. The BiCLIP baseline collapses with an accumulated OE ($0.1359$). This is most visible in the Oxford-Pet accuracy, which drops below zero-shot for BiCLIP ($72.7\%$) but remains at $86.3\%$ for GeoStack. Notably, GeoStack achieves a $+9.4\%$ lead in Average Accuracy across all tasks. While BiCLIP shows higher plasticity on a single task (Stanford Cars), it does so by sacrificing the integrity of all other experts. GeoStack’s ability to maintain an OE of $0.0142$—ten times lower than the baseline—validates its use as a scalable architecture for knowledge composition.

\section*{Appendix C: Stacking Metric.} 
\renewcommand{\thesubsection}{C.\arabic{subsection}}
\setcounter{subsection}{0}
\subsection{Orthogonality Error as a Metric for Stackability}
\label{Appendix:C}
\begin{figure}[ht]
    \centering
    \includegraphics[width=0.55\textwidth]{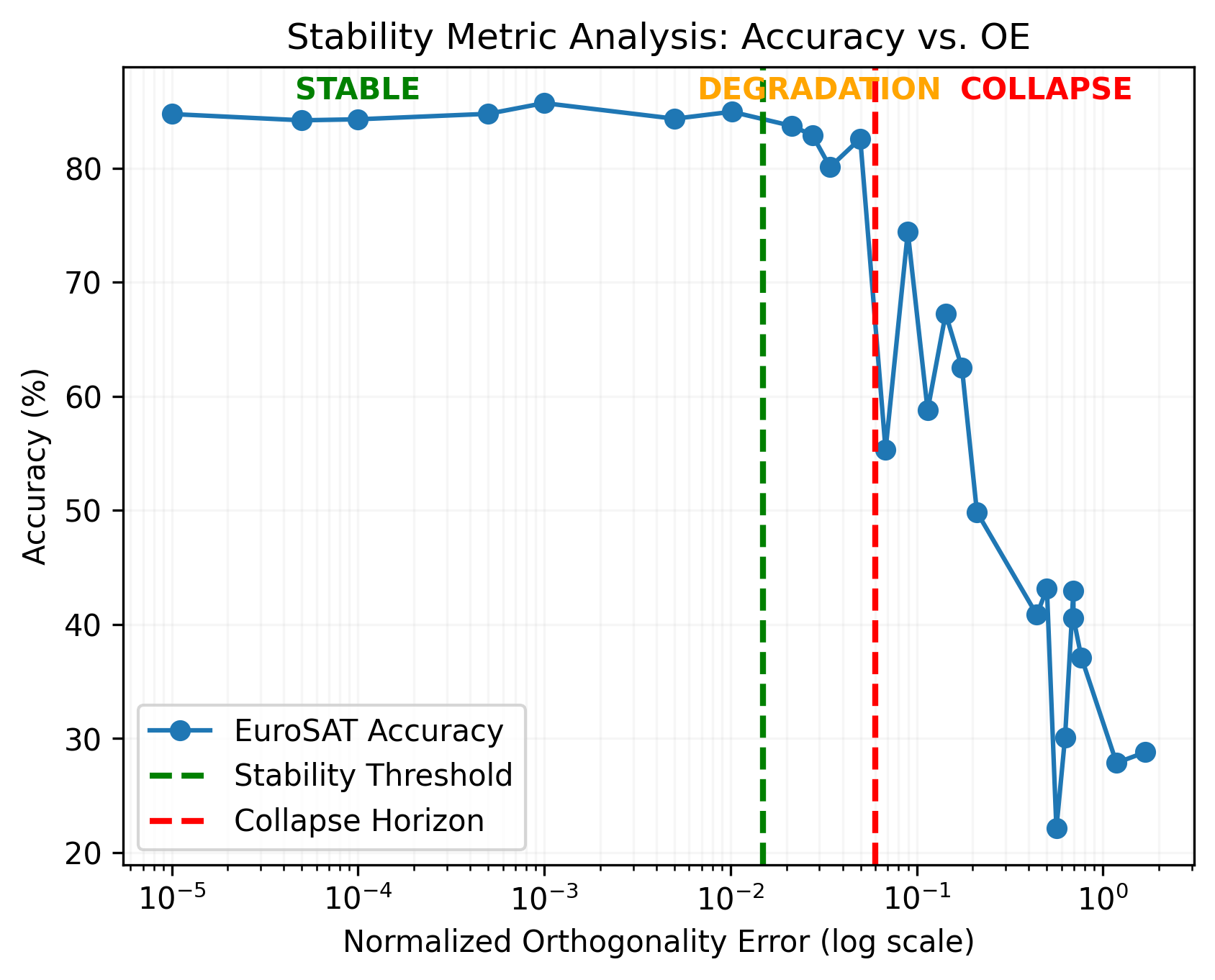}
    \caption{\textbf{Stability Stress Test Analysis.} EuroSAT accuracy as a function of simulated Orthogonality Error (OE). }
    \label{fig:stability_stress_test}
\end{figure}

In Sec.~\ref{sec:properties}, we defined the stackability metric as the normalized Orthogonality Error (OE): $$\mathcal{S}(W)^2 = \frac{1}{d^2} \| W^\top W - \mathbf{I} \|_F^2$$

Where $d$ is the model dimensions. Scaling by $1/d^2$ ensures that the error remains comparable even if the model's embedding dimension $d$ changes.

Here, we provide empirical results to calibrate this metric and identify the specific ranges that determine compatibility for weight folding. We conducted experiments by injecting synthetic experts with controlled OE. We synthesize matrices to match a target orthogonality error $\gamma \in [10^{-5}, 1.7]$. We folded these experts into a frozen backbone and evaluated the accuracy on the EuroSAT dataset.

The results shown in Fig.~\ref{fig:stability_stress_test} reveal three distinct ranges:

\begin{itemize}\item 
\textbf{The Stable Plateau ($\mathcal{S} < 0.015$):} Accuracy remains consistent in this range. The perturbation is small enough that the latent decision margins $M_a$ can absorb the interference without displacing previous knowledge.
\item \textbf{The Graceful Degradation Zone ($0.015 \leq \mathcal{S} < 0.06$):} The model retains most task-specific knowledge, the decision boundaries shift enough to cause a gradual decline in accuracy ($1\% \to 5\%$ drop).
\item \textbf{The Catastrophic Forgetting Horizon ($\mathcal{S} \geq 0.06$):} A threshold is crossed where the foundational knowledge is corrupted. This leads to a collapse of class separation, resulting in the rapid degradation in performance.
\end{itemize}

\section*{Appendix D: Broader Impacts.} 
\renewcommand{\thesubsection}{D.\arabic{subsection}}
\setcounter{subsection}{0}
\subsection{Positive Societal Impact}
\label{Appendix:D}
The primary positive impact of GeoStack is environmental sustainability. By enabling the folding of multiple domain experts into a single $O(1)$ inference operation, our method significantly reduces the computational energy and memory overhead required for multi-task deployments. 

\subsection{Negative Societal Impact \& Limitations}
As a foundational method for adapting pre-trained models like CLIP, GeoStack inherits the inherent biases and fairness issues present in the backbone architecture. GeoStack does not inherently audit or filter the semantic content of the experts being stacked. Consequently, an intentional misuse of the framework could involve the sequential stacking of harmful or biased experts.

\subsection{Mitigation}
Because GeoStack transformations are represented as transparent $d \times d$ linear layers rather than deep black-box adapters, they are more amenable to weight-space auditing.

\section*{Appendix E: Licenses.} 
\renewcommand{\thesubsection}{E.\arabic{subsection}}
\setcounter{subsection}{0}
\subsection{Assets}
\label{Appendix:E}

\begin{table}[h]
\centering
\caption{Asset Documentation: Versions, Licenses, and Access URLs.}
\label{tab:asset_details}
\small
\begin{tabular}{@{}llll@{}}
\toprule
\textbf{Asset} & \textbf{Version} & \textbf{License} & \textbf{Access / URL} \\ \midrule
OpenCLIP & ViT-B/16 & MIT & \href{https://github.com/mlfoundations/open_clip}{GitHub} \\
ImageNet & 2012 (1k) & Custom & \href{https://www.image-net.org/}{image-net.org} \\
CIFAR-100 & PyTorch/Torchvision & MIT & \href{https://www.cs.toronto.edu/~kriz/cifar.html}{cs.toronto.edu} \\
Caltech-101 & PyTorch/Torchvision & CC BY 4.0 & \href{https://data.caltech.edu/records/mzrjq-6wc02}{caltech.edu} \\
Flowers-102 & PyTorch/Torchvision & Custom & \href{https://www.robots.ox.ac.uk/~vgg/data/flowers/102/}{ox.ac.uk} \\
Food-101 & PyTorch/Torchvision & Custom & \href{https://data.vision.ee.ethz.ch/cvl/datasets_extra/food-101/}{ethz.ch} \\
DTD & v1.0 & CC BY-SA 4.0 & \href{https://www.robots.ox.ac.uk/~vgg/data/dtd/}{ox.ac.uk} \\
EuroSAT & RGB Version & MIT & \href{https://github.com/phelber/EuroSAT}{GitHub} \\
Stanford Cars & 2013 & Custom & \href{https://www.kaggle.com/datasets/eduardo4jesus/stanford-cars-dataset}{stanford.edu} \\
Oxford-Pets & v1.1 & CC BY-SA 4.0 & \href{https://www.robots.ox.ac.uk/~vgg/data/pets/}{ox.ac.uk} \\ \bottomrule
\end{tabular}
\end{table}

\section*{Appendix F: Compute Resources.} 
\renewcommand{\thesubsection}{F.\arabic{subsection}}
\setcounter{subsection}{0}
\subsection{Computational Resources and Efficiency}
\label{Appendix:F}
All experiments were conducted on a workstation equipped with an \textbf{NVIDIA GeForce RTX 2080 Ti (11GB VRAM)}. 

\textbf{Memory Usage:} During the training of a single GeoLayer adapter with a \textbf{ViT-B/16} backbone, peak memory usage was approximately \textbf{8.2 GB}. After weight folding, the inference memory footprint remains identical to that of the vanilla CLIP-ViT-B/16 projection head.

\textbf{Training Time:} Each domain-specific expert was trained for 10 to 30 epochs, with an average execution time of \textbf{25 minutes} per expert on a single NVIDIA RTX 2080 Ti.
% \newpage
% \input{checklist}
\end{document}